\documentclass[sigconf,natbib=false]{acmart}

\newcommand{\fig}{Figure }
\newcommand{\tab}{Table }
\newcommand{\eq}{equation}
\newcommand{\name}{InkStream}

\usepackage[normalem]{ulem}

\usepackage{amsmath,amsfonts} 
\usepackage{algorithmic}
\usepackage{graphicx}
\usepackage{textcomp}
\usepackage{xcolor}
\usepackage{multicol, multirow}
\usepackage[nocompress]{cite}

\setcopyright{none}
\renewcommand\footnotetextcopyrightpermission[1]{} 

\begin{document}

\title{
\name: Real-time GNN Inference
on Streaming Graphs via Incremental Update} 

\date{}

\author{Dan Wu}
\affiliation{%
  \institution{National University of Singapore}
  \city{Singapore}
  \country{Singapore}}
\email{danwu20@comp.nus.edu.sg}

\author{Zhaoying Li}
\affiliation{%
  \institution{National University of Singapore}
  \city{Singapore}
  \country{Singapore}
  }
\email{zhaoying@comp.nus.edu.sg}

\author{Tulika Mitra}
\affiliation{%
  \institution{National University of Singapore}
  \city{Singapore}
  \country{Singapore}
  }
\email{tulika@comp.nus.edu.sg}

\renewcommand{\shortauthors}{Wu et al.}

\thispagestyle{empty}

\begin{abstract}

Classic Graph Neural Network (GNN) inference approaches, designed for static graphs, are ill-suited for streaming graphs that evolve with time. The dynamism intrinsic to streaming graphs necessitates constant updates, posing unique challenges to acceleration on GPU. Inference on a $k$-layer GNN model, even after minor updates to the graph, requires accessing the entire $k$-hop neighborhood of the modified edges. Fetching the entire neighborhood is challenging as the GPU memory usually cannot accommodate the entire dataset. Thus the host needs to perform data fetch, impeding real-time inference. 

We address these challenges based on two key insights: (1) Inside the $k$-hop neighborhood,  a significant fraction of the nodes is not impacted by the modified edges when the model uses min or max as aggregation function; (2) When the model weights remain static while the graph structure changes, node embeddings can incrementally evolve over time by computing only the impacted part of the neighborhood.
With these insights, we propose a novel method, InkStream, designed for real-time inference with minimal memory access and computation, while ensuring an identical output to conventional methods. InkStream operates on the principle of propagating and fetching data only when necessary. It uses an event-based system to control inter-layer effect propagation and intra-layer incremental update of node embedding. InkStream is highly extensible and easily configurable by allowing users to create and process customized events. We showcase that less than 10 lines of additional user code are needed to support popular GNN models such as GCN, GraphSAGE, and GIN. Our experiments with three GNN models on four large graphs demonstrate that InkStream accelerates by 2.5-427$\times$ on a CPU cluster and 2.4-343$\times$ on two different GPU clusters while producing identical outputs as GNN model inference on the latest graph snapshot. 
\end{abstract}

\pagestyle{plain}
\maketitle

\section{Introduction}
Graph data structure is a versatile representation that consists of nodes interconnected by edges, enabling the modeling of complex relationships. Graph Neural Networks (GNN) are important tools in learning representations from graph-structured data. GNN takes a graph with each node holding a feature vector as input and generates an embedding vector for each node for the learned node representation. The node embedding can be then used for downstream tasks, such as link prediction in social networks~\cite{survey_recommend}, traffic forecasting in road networks~\cite{survey_traffic}, molecule classification~\cite{survey_molecular}, etc. A GNN model consists of multiple GNN layers. In each layer, every node collects information from its direct neighbors and updates its embedding.

The prevalent usage of GNNs has mostly revolved around static graphs, overlooking the dynamic nature of many real-world graphs \cite{dygnn, pareja2020evolvegcn, huan2022t}. Such graphs, referred to as \textit{streaming graphs}, evolve over time, requiring frequent update of node representations~\cite{chang2017streaming, yang2022streaming, bielak2022fildne}. Social networks like Twitter or Facebook are classic examples where new users join, existing ones leave, and relationships form and cease continuously. E-commerce platforms are other excellent examples, where new users and items are continually added, and user-item interactions are constantly updated. 

Streaming graphs pose a challenge to graph learning: the learned node embedding trained on the offline static datasets cannot reflect the current status. Several works adapt the node embedding to the evolved graph structure by frequent model retraining \cite{pareja2020evolvegcn, dygnn, huan2022t, he2023dynamically}. However, in numerous real-world applications such as recommendation systems, there is a need for prompt, real-time responses  \cite{chang2017streaming, yang2022streaming, huan2022t}. This renders the extended retraining periods of these methods impractical. Striking a balance between maintaining model accuracy and ensuring swift responses is essential. A practical approach is to periodically retrain the model at fixed intervals on a dedicated server, while using a snapshot of the model for instant node embedding update when the graph structure changes during the retraining interval. This tactic offers the advantage of quick responses while concurrently aligning with the evolving graph structure, thereby maintaining the freshness of the graph data.

\begin{figure}[t] 
\centering
\includegraphics[width=\columnwidth]{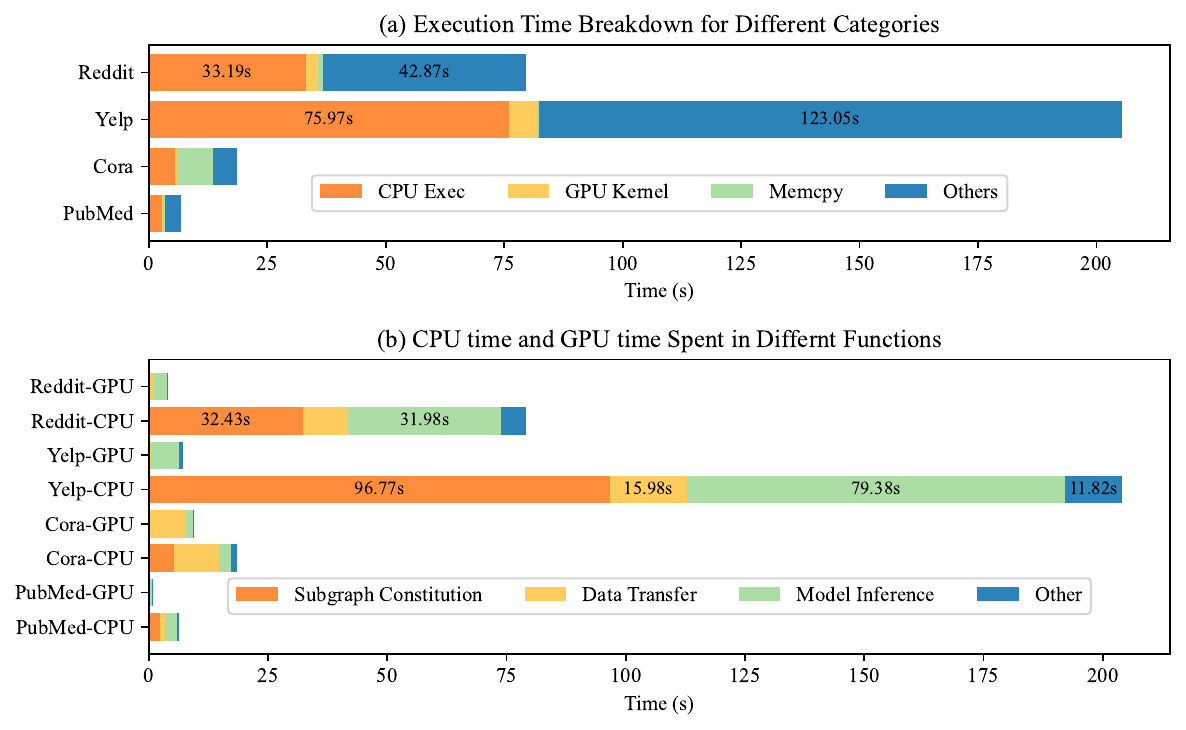}
\caption{Execution time breakdown in 2-layer GCN. Off-chip memory access is included in Others. Memcpy refers to GPU-involved memory copy time.}
\label{fig:time_breakdown}
\end{figure}

Unfortunately, classical approach to GNN inference on an evolving graph cannot meet the real-time requirement due to extensive memory access. Real-world datasets can easily reach up to GB or TB scale and cannot fully fit into the GPU memory. Such datasets  can be handled via mini-batch inference \cite{li2022hyperscale, survey_sampling, zhong2023cognn, hamilton2017inductive}. Here, the first step is to construct a subgraph consisting of $k$-hop neighborhoods of selected nodes, e.g., nodes with consecutive ids. This can only be done on the host device (CPU in most cases), which is time-consuming, involving significant memory access and CPU-GPU communication \cite{li2022hyperscale}. 
\fig\ref{fig:time_breakdown} shows the execution time breakdown of a 2-layer Graph Convolution Network\cite{gcn} on a CPU-GPU platform. As shown in \fig\ref{fig:time_breakdown}(a), the CPU execution and others (off-chip memory access is categorized as others) take most of the time, while GPU kernels and GPU memory copy occupy only a small fraction of the total runtime.
After the subgraph constitution, the graph structure and node features are sent to the GPU for computation. Finally, the computed node embeddings are sent back to the CPU. \fig\ref{fig:time_breakdown}(b) shows the CPU and GPU time spent on different functions during inference, namely, subgraph construction, data transfer between CPU and GPU, and the computation of model inference. The subgraph construction takes a significant portion of the execution time. 

\begin{figure}[t] 
\centering
\includegraphics[width=\columnwidth, trim={0 0.2cm 0 0},clip]{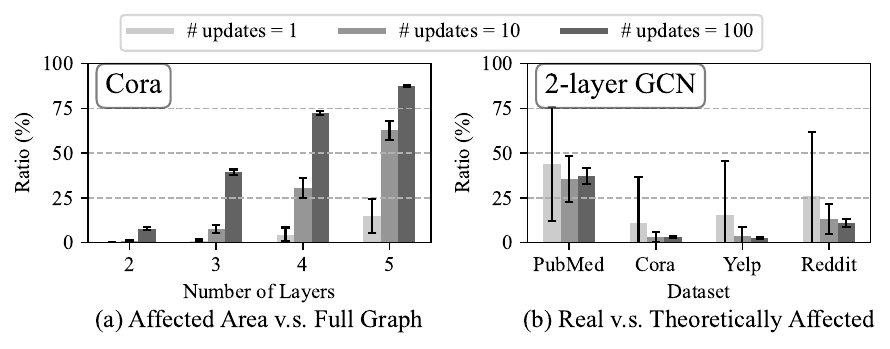}
\caption{Ratio of the theoretically affected area on full graph, and the ratio of real affected nodes in the theoretically affected area.}
\label{figure:motivating_data}
\end{figure}

As a node can only get information from its direct neighbors in a GNN layer, a node can be affected by nodes in a $k$-hop neighborhood with a $k$-layer GNN model. Therefore, the $k$-hop neighborhoods of source nodes of changed edges constitute an \textit{affected area}. 
\fig\ref{figure:motivating_data}(a) shows the ratio of the affected area of "\# updates" changed edges, compared with the whole graph for GNNs of different numbers of layers. It can be seen that the affected area can be tiny (less than 1\%), compared with the whole graph, when $k$ is small.
A straightforward way to take advantage of this insight is to only compute for the nodes in an affected area\cite{schaefer2022aegnn, dygnn}. Unfortunately, this still takes seconds with the insertion of merely one edge.

\begin{figure}[t] 
\centering
\includegraphics[width=\columnwidth, trim={0 0.3cm 0 0},clip]{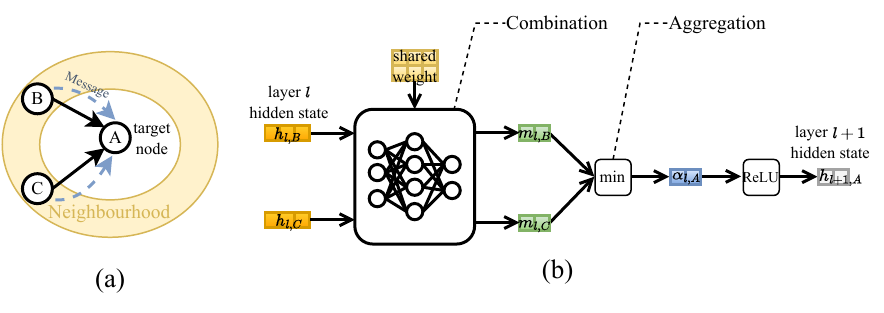}
\caption{(a) Neighborhood message-passing mechanism of a GNN layer. (b) Computing abstraction of message-passing mechanism.}
\label{fig:message_passing}
\end{figure}

We observe that when the model weights remain static while the graph structure changes, \textit{a considerable proportion of nodes in the theoretically affected area are actually not impacted by the changed edges when the model uses monotonic aggregation function}, like min and max. 
To explain the underlying reason, we first briefly introduce the general message-passing framework adopted by most GNN models. In a multi-layer GNN model, each layer generates a hidden state,  an intermediate node embedding vector. A GNN layer can be expressed in the message-passing framework by  two primary execution phases: \textit{Combination} and \textit{Aggregation} \cite{mpnn, yan2020hygcn}, as shown in  \fig \ref{fig:message_passing}.
The combination phase acts like neural networks, where each node transforms its embedding with a shared multi-layer perceptron. In the aggregation phase, each node receives its neighboring nodes' embeddings, then reduces them into a single vector with a selected aggregation function. The aggregation function is invariant to the permutations of node orderings, such as a mean, sum, or max function \cite{survey_gnn}. 
Especially when the min or max function is chosen, the aggregation is selective and monotonic, i.e., a node does not contribute to a neighbor if it is not selected  as min or max. \fig\ref{figure:motivating_data}(b) shows the ratio of real affected nodes versus the theoretically affected area. On average, only 3\% nodes are influenced in Cora and Yelp datasets with 100 updates. 

Despite the existence of unimpacted nodes in the affected area, the classical method cannot directly bypass their computation. The reason is two-fold. First, such methods target static graphs, running inference on the graph structure at a specific timestamp without knowledge of the results in the previous timestamp. Second, even if a previous result is given, these methods fail to identify and skip the computation for uninfluenced nodes because it preselects the node set before any computation is invoked when constructing the subgraph.

We also notice that \textit{when the model stays static, the node embedding can be incrementally updated in a layer by only fetching and computing for the affected neighbors instead of the whole neighborhood}. 
Incremental update stems from the observation that node embeddings at a given timestamp $t$ encapsulate the information of the neighborhood at time $t$. As these neighborhoods often show minimal changes in a short time, the transition from timestamp $t$ to $t$+1 only slightly alters the neighborhood. Consequently, it is feasible and efficient to reuse the embeddings at timestamp $t$ and bypass the computation of the unchanged portion of the neighborhood, thereby mitigating computational requirements. To do so, a node must first cancel the impact for a changed neighbor's old embedding, then add the impact of the new embedding. 

\begin{figure*}[t]
\centering
\includegraphics[width=6.5in, trim={0 0.2cm 0 0},clip]{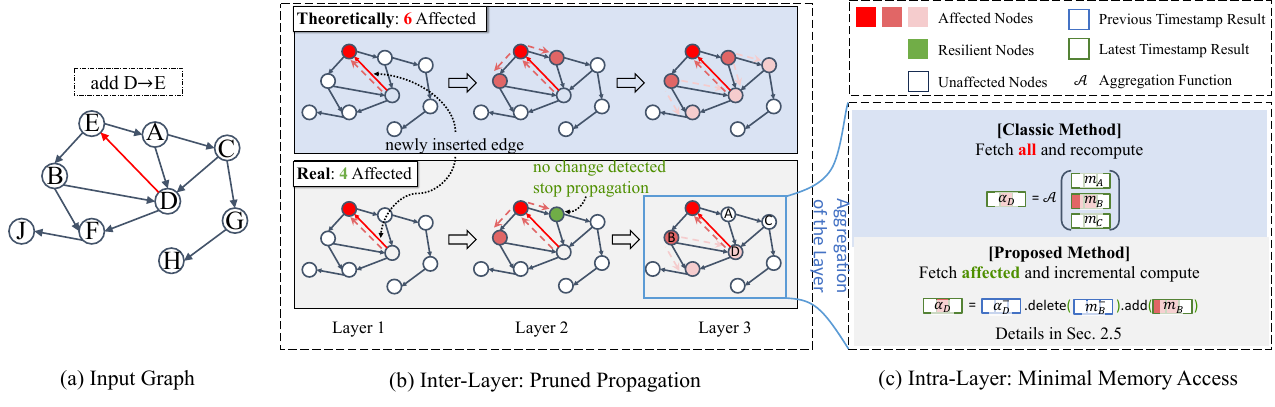}
\caption[overview]{Two-level savings in \name. $\mathcal{A}$ for aggregation function. $m$ and  $\alpha$ are node embedding before and after aggregation. Superscript~$^-$ denotes the result in the previous timestamp.}
\label{figure:overview}
\end{figure*}

Impact cancellation with a monotonic function is not always safe as it would lead to irrecoverable data loss . Suppose a node has a high-impact neighbor that contributes to most of the node's embedding, making it insensitive to the rest of the neighborhood. Then canceling the influence of this high-impact neighbor would lead to irrecoverable data loss, and a node need to recalculate by fetching its whole neighborhood. 

To fully exploit the aforementioned two findings, we propose a novel and generic method called \name, addressing the ensuing challenges. \name\ allows real-time GNN inference with minimal memory access and computation while enjoying zero accuracy loss compared with conventional method of inference on the graph's latest snapshot.

\name's design principle is "\textit{Propagate only when necessary. Fetch only the necessary.}" 
\textbf{First}, in a GNN model, a changed embedding of one node at a layer will affect its immediate neighbors' embeddings at the next layer, thus forming a propagation tree. We prune the propagation tree to avoid unnecessary computation.  Suppose there is an edge inserted from node D to E, as shown in \fig\ref{figure:overview}(a). Across the layers, \name\ mitigates unnecessary computation by identifying \textit{resilient node}s (green node in the figure) in the propagation tree and pruning the corresponding subtrees rooted at these nodes (\fig\ref{figure:overview}(b)). A node is resilient if it could be affected but turns out to be uninfluenced.  We design an \textit{event}-based approach to control the propagation. An event is created and sent along the edges when a node changes its embedding, and vice versa.
 \textbf{Second}, inside each layer, during the aggregation phase, \name\ reuses data from the previous timestamp, incrementally applying the effects of changed neighbors. 
 As shown in Figure~\ref{figure:overview}(c), \name\ deletes the impact of changed neighbors' old information and adds the impact for new information, thus incrementally evolving to the latest node embedding. By only accessing prior results and affected neighbors' information without fetching the whole neighborhood, \name\ drastically minimizes memory access and execution time. 
 
While saving computation and memory access, \name\ also leverages the selective features of monotonic aggregation functions to ensure identical results as classical methods.
This is attributed to the nature of monotonic functions, where the final result can be seen as a copy of the selected input. Once the same selection is made, identical results can be ensured. 
Moreover, \name\ has good extensibility by providing interfaces to allow the users to easily configure for their own model. On three test cases: GCN\cite{gcn}, GraphSAGE\cite{hamilton2017inductive}, and GIN\cite{gin}, users only need to provide less than 10 lines of additional code.  In a nutshell, \name\ is a highly extensible method that minimizes the computation and memory access at the inter-layer level and intra-layer level while bringing no accuracy loss.

\begin{figure}[t] 
\centering
\includegraphics[width=0.9\columnwidth]{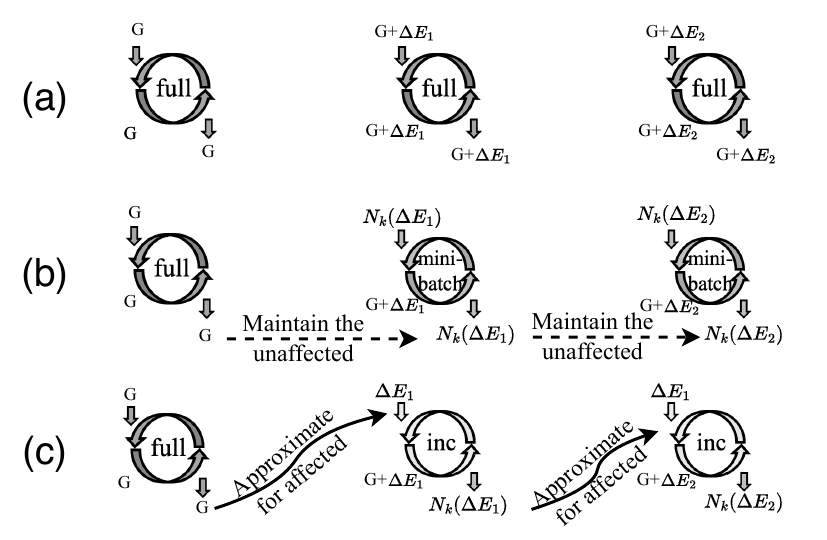}
\caption{Inference on a streaming graph using (a) classic method, (b) existing method, and (c) proposed method. 
$G$ is the graph. $\Delta E$ denotes the changed edges between two timestamps. $N_k(\Delta E)$ is the union of $k$-hop neighborhoods of source nodes of edges in $\Delta E$.
}
\label{fig:streaing_graph_inference}
\end{figure}

 A visual comparison of the classic method, the existing method, and the proposed method is shown in \fig \ref{fig:streaing_graph_inference}. In the classic method, the graph structure is first updated, then inference on the full graph is performed to get the updated node embedding. For the existing method \cite{dygnn, schaefer2022aegnn}, only the embeddings for affected nodes are updated through recompute, while the rest remain the same. We took one step further: not only the embeddings of unaffected nodes are preserved, but the previous state embeddings of affected nodes are also reused to calculate the updated ones.

Our contributions can be summarized as follows:
\begin{itemize}
\item \textbf{New Findings}: We found the gap between the theoretically affected area and the real affected area for GNN inference on streaming graphs with a monotonic aggregator. We also found that a node embedding can evolve incrementally from an old result in the previous timestamp.
\item \textbf{Minimal Overhead}: \name\ requires minimal data access and computation by pruning the effect propagation tree and reusing the previous results. 
\item \textbf{Generality}: Our method can be applied to a wide range of GNNs through simple configuration.
\item \textbf{No Accuracy Loss}: \name\ generates identical results as the conventional method and brings no loss to the model accuracy.
\item \textbf{Large Performance Improvement}: Evaluated on three GNN models and four datasets, experiments show a 2.5-427$\times$ speedup on a CPU cluster and 2.4-343$\times$ speedup on two different GPU clusters.
\end{itemize}

\section{\name}

\name\ takes the result of an initial full graph inference as input and incrementally updates with streaming edges. During the full graph inference, we save the embedding before and after aggregation for the whole node set in all layers. Since the hidden state length is usually carefully designed and small \cite{yoo2023sgcn}, the memory cost is acceptable and worth the speedup. For clarity, we first introduce the notations used in this paper and the expressiveness of \name. 
Later, we present the overall workflow and each part of the workflow in detail.

\subsection{Notations for Streaming Graphs}
\tab \ref{table:notation} lists the notations used in the paper for streaming graphs and GNNs. A graph is denoted as $G$($V$,$E$), with vertex set $V$ and edge set $E$. Between two timestamps, a set of edges $\Delta E$ is modified,  either inserted or removed. The direct neighborhood of a vertex $u$ is $N(u)$.
In the message-passing mechanism, there is a combination phase and aggregation phase, with flexible ordering. The operations in each phase can also be customized for different tasks. We abstract the computation in combination phase into a function $\mathcal{T}()$, and the aggregation function $\mathcal{A}()$. 
At the end of each layer, there is an element-wise activation function $act()$.
In a multi-layer GNN model, we name the embedding of node $u$ before being fed into layer $l$ as $h_{l,u}$, which is the input of layer $l$. 
Inside the layer $l$, we name the input and output of the aggregation phase as the \textbf{message} $m_l$ and the \textbf{aggregated neighborhood} $\alpha_l$ respectively, where $\alpha_{l,u} = \mathcal{A}(m_{l,v}: v\in{N(u)})$.
When aggregation phase is first executed in a layer, $m_{l} = h_{l}$, otherwise, $m_l = \mathcal{T}(h_l)$. 
To distinguish results in previous and current timestamps, a superscript $^-$ is used. Any symbol with a superscript $^-$ refers to the one in the previous timestamp and vice versa.

\begin{table}[h] \footnotesize 
\caption{Notations}
  \label{table:notation}
  \centering
\begin{tabular}{ | c | l | }
 \hline 
\textbf{Notations} & \textbf{Descriptions} \\ \hline
$G$($V$,$E$) & graph $G$ with vertices $V$ and edges $E$\\ \hline
$\Delta E$ & modified edges between two timestamps\\\hline
$N$($u$) & direct neighborhood of vertex $u$ \\ 
\hline \hline

$\mathcal{A()}$ & aggregation function
\\\hline
$\mathcal{T}()$ & combination function\\\hline
$act()$& activation function\\\hline\hline

$h_{l,u}$ &  hidden state of vertex $u$ in layer $l$\\ \hline
$m_{l,u}$ & message sent from vertex $u$ in layer $l$  \\ \hline
$\alpha_{l,u}$ & embedding of $u$ after aggregation in layer $l$ \\ 
\hline\hline

$\Box$ & data in current timestamp \\\hline
$\Box^{-}$ & data in previous timestamp \\\hline
\end{tabular}
\end{table}
\subsection{Expressiveness}
\name\ is capable of describing GNNs built on the message-passing framework with two certain characteristics:

1. The aggregation phase act as a \textit{selection} from direct neighbors: \newline $\alpha_{l, u}=\mathcal{A}({m_{l,v}:v\in N(u)}),$ $\mathcal{A}=min/max$. The selective characteristics of aggregation function $\mathcal{A}$ is the cornerstone of \name, where computation can be pruned at the layer level, and incremental updates bring zero loss. Min and max are commonly-used aggregation functions and show competitive expressiveness against other aggregation functions across varied GNN models \cite{pellegrini2020learning, hamilton2017inductive, pna, dgcnn}. 

2. One node's message in layer $l$+1 \textit{only} depends on its message and aggregated neighborhood in layer $l$: $m_{l+1, u} = \mathcal{T}(\alpha_{l, u}, m_{l,u})$. In this case, the effect of a changed edge can only propagate along the edges or to the node itself in the next layer. Thus, the changed edge will not influence a significant portion of the graph.
This also means \name\ cannot support specific operations, like batch normalization, graph normalization, and aggregation with normalized weight.
One example is Graph Attention Networks (GAT)\cite{gat} with attention coefficients between two nodes calculated by:
\begin{equation}
 a_{u v}=\frac{\exp \left(\operatorname{LeakyReLU}\left(\mathbf{a}\left[\mathbf{W} h_u \| \mathbf{W} h_v\right]\right)\right)}{\sum_{k \in N_u} \exp \left(\operatorname{LeakyReLU}\left(\mathbf{a}\left[\mathbf{W} h_u \| \mathbf{W} h_k\right]\right)\right)}
\end{equation}
where $\|$ is the concatenation operation, $\mathbf{a}$ and $\mathbf{W}$ are weight vector and weight matrix.

A detailed showcase will be presented in section \ref{sec:case_study}, introducing how different GNN models can be supported with \name.

\subsection{InkStream Workflow}
For a multi-layer GNN model, Inkstream creates initial $event$s for changed edges when processing the first layer.
An \textit{event} is a message carrying the operation, target node, and an embedding vector. It tells whether to add or cancel the impact of the vector on the target node's embedding. Then in subsequent layers, each changed node creates and propagates the events to neighbors, to be processed in the next layer. 
Inside each layer, nodes with at least one event heading to it will be iteratively visited and updated.
To be more specific, at the beginning of each layer, events are created and pushed to the current layer's event queue for each changed edge (section~\ref{sec:propagation}).
Then, all events heading to the same target node with the same operation are grouped and reduced into one single event (section~\ref{sec:grouping}). Each time, all reduced events with the same target node are popped out from the event list, with no more than one reduced event for each operation. \name\ will check whether incremental update can be applied to the target node (section~\ref{sec:evolvable_condition}). If applicable, the incremental update is executed (section~\ref{sec:incremental_update}). Otherwise, information of all neighbors is fetched, and the target node's embedding is recomputed in a classic method. For nodes incrementally updated, there is a chance of resilience to the changes carried by events, so \name\ checks whether the node embedding has changed. If so, it will not propagate any event to its neighbors, which means the propagation stops early. In another case, where node embedding changes and is thereafter transformed with operations in the combination phase, this node creates events for its neighbors and pushes the events to the next layer's event queue (section~\ref{sec:incremental_update}). Specifically, this node propagates events along the edges in the previous timestamp to cancel the impact of its old node embedding, then propagates events to neighbors in the current timestamp to add the impact of its new embedding.

\begin{figure}[t] 
\centering
\includegraphics[width=0.6\columnwidth, trim={0 0 0 0},clip]{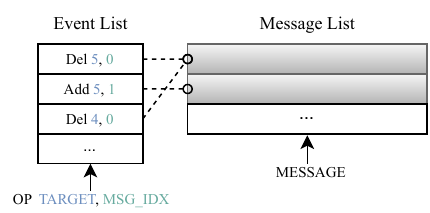}
\caption{Illustration of \textit{event queue}, containing an event list and a message list.}
\label{fig:event_queue}
\end{figure}

\subsection{Inter-Layer: Event-based Computing Model}
\begin{figure}[t] 
\centering
\includegraphics[width=\columnwidth, trim={0.3cm 0.3cm 0.3cm 0},clip]{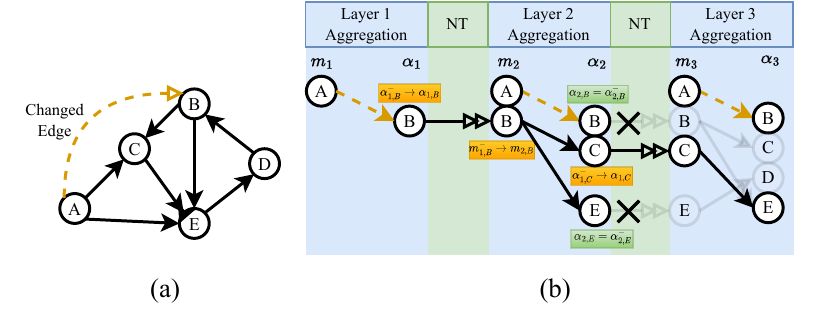}
\caption{Layer-level event propagation. NT stands for node transformation, referring to the operations taking place between the aggregation phases of two consecutive GNN layers, containing combination and activation.}
\label{fig:propagation}
\end{figure}

We proposed the \textit{event}-based computing model to manage the inter-layer propagation. An event is a message carrying the operation, target node, and embedding vector.
The embedding vector is heavy and shared among multiple events (e.g., a node tells all its neighbors to add an impact of its new embedding). 
Therefore, we use the \textit{event queue} to efficiently manage the events containing an \textit{event list} storing the metadata, and a \textit{message list} accommodating the shared node embedding information in events, as shown in \fig \ref{fig:event_queue}. 
For example, \{\texttt{Del}, 5, 0\} means to cancel the contribution of vector stored at index 0 in the \textit{message list} to the embedding of node 5 in the current layer. 
With the help of our event-based computing model, we can easily manage the computation for the affected area. An illustrative example is shown in \fig \ref{fig:propagation}. If one node's embedding remains the same after finishing the impact addition and cancellation in the events heading to it, it will not further propagate events to save computation afterward. The event-based computing model has two primary components: 
1) {Grouping}: group and reduce events heading to the same target node.
2) {Propagation}: propagate the effect to neighbors. 

\begin{figure}[t] 
\centering
\includegraphics[width=\columnwidth, trim={0.3cm 0.5cm 0.3cm 0.1cm},clip]{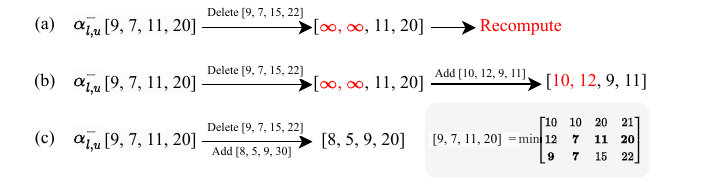}
\caption{Illustrative example for the necessity of grouping. (a) Recomputation is costly but correct. (b) Incremental updates for each event independently lead to a wrong result. (c) Analyzing all events to decide whether efficient incremental updates can be applied ensures a correct result while saving computation.}
\label{fig:reduce_example}
\end{figure}

\subsubsection{Event Grouping}
\label{sec:grouping}
\textbf{Necessity of event grouping.} 
If one node $u$ has multiple neighbors changed in the former layer $l$-1, it will receive several events, non-continuously stored in the event list. It is inefficient to iterate the event list sequentially and process each event separately as the old aggregated neighborhood needs to be fetched each time. Instead, we group the events heading to the same node and process all of them at the same time. Apart from repeated memory access, event grouping also improves the evolvability of nodes. For example, as shown in \fig \ref{fig:reduce_example}, one target node has two events from different source nodes, one for impact deletion and the other for addition. After the deletion, the first half of the embedding is found affected and needs to be \textit{reset} with a default value ($\infty$ if $\mathcal{A}$=min, otherwise $-\infty$). If the two events are processed independently, since there are two irrecoverable elements after deletion, we need to fetch all neighbors' information and recompute in a classic method to get the correct result. For another option, if the incremental update is directly applied without knowing other events, a wrong result could be derived. However, if all events are processed together, we can decide whether the efficient incremental update can be applied to ensure a correct result with minimum effort. 
After grouping by target node, the events with the same operation are reduced into one event to reduce data movement in the following process. The feasibility of reduction will be explained in section \ref{sec:evolvable_condition}.

\noindent\textbf{Process of event grouping.} 
To conduct the event grouping, we iterate through the event list and group the events by first the target node, then the operation. 
Next, we reduce all grouped events in the same group with the aggregation function.

\subsubsection{Event Propagation}
\label{sec:propagation}
 When one node has 1) its embedding or 2) connectivity changed in a layer, a set of events will be created for each neighbor, carrying the message whose impact will be canceled or added. Then these events will be pushed to the event queue for the next layer. In this section, we introduce how these two conditions are handled in our event-based computing model.

\noindent \textbf{Propagate for changed edges.}
At the \textit{beginning} of processing one layer $l$, we create events for destinations of changed edges. The events will later be consumed in the processing of this layer.
Suppose an edge ($u$, $v$) is removed. The old message $m_{l, u}^-$ will first be pushed to the message list and returns a pointer to this message, msg\_idx. Then, an event containing the operation \texttt{Del},  the target node $v$, and the pointer msg\_idx will be created. For inserted edge ($s$, $t$), the new message (remains unaffected or updated when processing former layer) $m_{l, s}$  is pushed, with an \texttt{Add} operation. 

\noindent \textbf{Propagate for affected nodes.}
 At the \textit{end} of processing the layer $l$-1, we create events for the neighbors of those affected nodes in the next layer $l$. These events will later be consumed in the processing of the next layer. The process is different from how changed edges are processed because the impact of the old message needs to be canceled first before adding the new one. For an affected node $u$ whose aggregated neighborhood changed from $\alpha_{l, u}^-$ to $\alpha_{l, u}$, we first calculate its message in the next layer according to the model configuration, like $m_{l+1,u} = act(\mathcal{T}(\alpha_{l,u}))$. Then messages in two states, $m_{l+1, u}^-$ and $m_{l+1, u}$, are pushed to the message list, given two indices $idx_u^-$ and $idx_u$. We iterate though its neighborhood in previous timestamp $N(u)^-$, and create an event (\texttt{Del}, $v$, $idx_u^-$) for impact cancellation for $v\in N^-(u)$. Lastly, we create the event (\texttt{Add}, $v$,  $idx_u$) for impact addition for neighbors in current timestamp, $v\in N(u)$.
 
Duplicated events exist if one node is the destination of a changed edge meanwhile the source node is affected in the former layer. However, the duplication will not affect the final result as the messages in events will be fed into a monotonic aggregation function. The memory overhead caused by this duplication is also trivial considering the ratio of changed edges and affected area.

\subsection{Intra-layer: Incremental Update}
Inside each layer, after event grouping, \name\ takes the grouped events of a target node, checks the evolvability, and either applies incremental update or recomputes the embedding in a classic way. In the section, we first introduce the applicable conditions of incremental update, then the process of incremental update and recomputing.

\subsubsection{Applicable Conditions}
\label{sec:evolvable_condition}
The effect of events on the old aggregated neighborhood can fall into three categories, and incremental updates can be applied in two of them.
For clarity, we define the notations first.
After event grouping with no reduction, target node $u$ has a set of messages to be deleted [$m^-_1, m^-_2, \ldots, m^-_p$] and a set of messages to be added [$m_1, m_2, \ldots, m_q$].
Especially for target node's old result, $\alpha_{l,u}^-$, if there is any index where the old result equals to the message to be deleted, $\exists\, i\, \in\, [0,\, \vert \alpha_{l,u}^- \vert], j\in\,[1,\,p]$, such that $\alpha_{l,u}^-[i]=m^-_j[i]$, then $\alpha_{l,u}^-$ need to be \textit{reset} at position~$i$. 
Since the aggregation function is monotonic and the messages to be deleted are a subset of the whole neighborhood information $\alpha_{l,u}^-$ encapsulates, a reset position can only appear in the message whose value at that index is the minimum (maximum) among all messages to be deleted. Therefore, without loss of correctness, these messages are reduced before checking: $m_\mathcal{A}^- = \mathcal{A}(m^-_1, m^-_2, \ldots , m^-_p)$,  $m_\mathcal{A} = \mathcal{A}(m_1, m_2, \ldots , m_q)$.

\begin{figure}[t] 
\centering
\includegraphics[width=\columnwidth, trim={1cm 0 0 0},clip ] {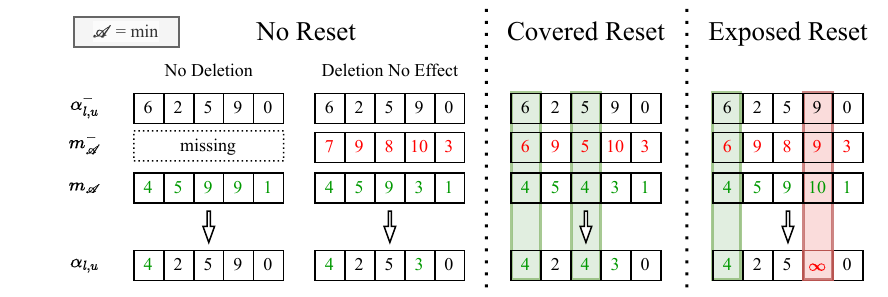}
\caption{Events' Effect: Three Conditions.}
\label{fig:applicable_conditions}
\end{figure}

There are three conditions of how events affect the old result: \textit{no reset}, \textit{covered reset}, and \textit{exposed reset}. An illustrative example is shown in \fig \ref{fig:applicable_conditions}.

\noindent \textbf{No Reset.} When there is no position that needs to be reset, $\forall\, i \, \in \, [0, \, \vert \alpha_{l,u}^- \vert],\ \alpha_{l,u}^-[i] \neq m_\mathcal{A}^-[i]$, incremental update can be applied. 
Especially there are two cases in this condition: 1) there is no message to be deleted; 2) the deletion has no influence on the old result $\alpha_{l,u}^-$. 

\noindent \textbf{Covered Reset.} When the positions that need to be reset are \textit{cover}ed by the message to be added, incremental updates can also be applied. Suppose there is a set of positions that need to be reset, $D$. We say $D$ is covered by the new message $m_\mathcal{A}$, if for any position $i$ in $D$, the message to be added is 'better' than the message to be deleted, $\mathcal{A}(m_\mathcal{A}^-[i], m_\mathcal{A}[i]) = m_\mathcal{A}[i]$. 
Despite the fact that part of the old result is affected and needs to be reset, the correctness of incremental updates can be ensured due to the transitivity monotonic function.
The key insight is the fact that messages from unaffected neighbors are no better than the deleted value. Therefore if the reduced new message is better than the deleted value, it must be better than others. 

\noindent \textbf{Exposed Reset.}
When positions need to be reset and are not fully covered by the new message, recompute is necessary to ensure a correct result.

\begin{figure*}[t]
\centering
\includegraphics[width=7.2in]{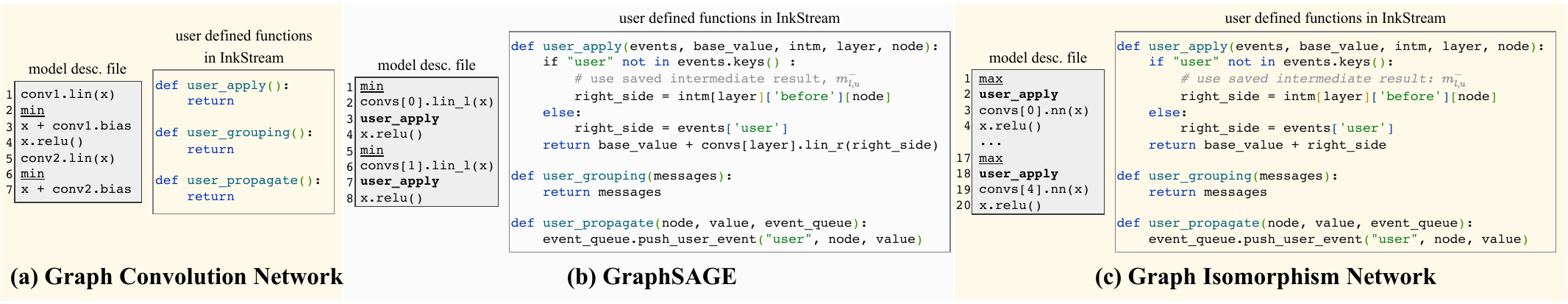}
\caption[configurations]{Examples of additional code and configuration files to support different GNN models. In the model description file, each line corresponds to an operation. Keywords \underline{min} and \underline{max} indicate incremental updates for the aggregation phase with the selected aggregation function. Keyword \textbf{user\_apply} tells \name\ to execute {\fontfamily{qcr}\selectfont user\_apply} function.  }
\label{figure:showcase_configs}
\end{figure*}
\subsubsection{Update and Propagate}
\label{sec:incremental_update}
Once incremental update can be applied to a node $u$, it will be updated with the reduced addition messages: $\alpha_{l,u} = \mathcal{A}( \alpha_{l,u}^-, m_\mathcal{A})$.
Specifically, in \textit{no reset} condition, there is also a chance that the new message has no impact on the old result, such that $\alpha_{l,u} = \alpha_{l,u}^-$. Therefore, we compare $\alpha_{l,u}$ against $\alpha_{l,u}^-$ to check whether the node has evolved. If the node is resilient to the changes, it will stop propagation. In other cases, where node 1) has been updated without a reset, 2) or has covered reset, 3) or is recomputed, inter-layer propagation will be initiated. 

When the incremental update is not applicable,  all messages from neighbors in the current timestamp are fetched and aggregated to recompute the node: $\alpha_{l,u} = \mathcal{A}(m_{l,v}), \ v \in N(u)$. Note that no additional computation is required to get $m_{l,v}$. As $m_{l,v}$ is calculated from $\alpha_{l-1,v}$,  $m_{l,v} \neq m^-_{l,v}$ if and only if  $\alpha_{l-1,v} \neq \alpha^-_{l-1,v}$.
There are only three types of nodes in a layer: unaffected, resilient, and affected. If $v$ is unaffected or resilient in layer $l-1$, then $\alpha_{l-1,v} = \alpha^-_{l-1,v}$, $m_{l,v} = m^-_{l,v}$. If $v$ is affected in layer $l-1$, $\alpha_{l-1,v}$ and $m_{l,v}$ will be computed and updated when processing layer $l-1$.

\subsection{Flexible User-Defined Functions}
To support models with more complex operations, we allow users to additionally define their own event creation, propagation, grouping, and apply functions apart from the management of native events in \name. 
Specifically, \name\ provides three interfaces for customization: {\fontfamily{qcr}\selectfont user\_propagate}, {\fontfamily{qcr}\selectfont user\_grouping} and {\fontfamily{qcr}\selectfont user\_apply} functions. {\fontfamily{qcr}\selectfont user\_propagate} allows the user to create and propagate customized events, executed together with \name's built-in propagation function. {\fontfamily{qcr}\selectfont user\_grouping} defines the grouping and processing of customized events during the events grouping process. {\fontfamily{qcr}\selectfont user\_apply} contains the logic to apply the effect of customized events to the tensor at a phase level. The execution order of different phases can be flexibly managed by providing a short model configuration, as shown in \fig\ref{figure:showcase_configs}. The management of user-defined events is isolated from the native functions.

\subsection{Support for other operators}
\noindent\textbf{Sampling.}
Graph sampling is an important technique for GNN training to solve the problem that the graph and intermediate embeddings cannot fit into GPU memory. Since \name\ is GPU-memory-friendly, no sampling is used in \name\ by default. However, sampling operations can be supported by propagating events only to selected neighbors.
Regardless of the sampling strategy, one simple solution is to cache and reuse previous samples for streaming graphs \cite{sketching}, while for newly inserted edges, we assume it is always selected.

\noindent\textbf{Pooling.}
\name\ is an event-based method manipulated at the node level, so the graph-level pooling operation cannot be natively supported in InkStream. Instead, \name\ handles the pooling operations in its original way: after updating the node embedding for layers before the pooling layer, inputs for the pooling layer will be fetched, then the pooling operation is executed. 

\section{Integration with GNN models}
\label{sec:case_study}

In this section, we detail how \name\ supports various GNN models using a high-level model description file and three user-defined functions, as depicted in \fig \ref{figure:showcase_configs}. Within the model description file, each line represents an operation to be executed sequentially. There are two primary keywords: min/max and user\_apply. Lines containing min or max keywords denote executing the incremental update for the aggregation phase, while those with user\_apply refer to call the {\fontfamily{qcr}\selectfont user\_apply} function. Instead of the nature division of layers, we partition the model configuration by aggregation indicators, as the events are propagated right before the aggregation and reduced during the aggregation phase. For example, GCN's model description in \fig \ref{figure:showcase_configs}(a) will be partitioned into three parts, starting from lines 1, 2, and 6 separately. The event propagation will be conducted at the end of each partition. Then events are grouped at the start of the aggregation phase in the next partition. 

\subsection{GCN}
Graph convolution network (GCN) \cite{gcn} is one of the most successful GNNs for graph learning, bridging the gap between spectral-based approaches and spatial-based approaches\cite{survey_gnn}. The mathematical expression of one GCN layer is:
\begin{align}
    & a_{l,v} = \mathcal{A}(\{h_{l-1,u}: u\in N(v) \}), \label{eq:1} \\
    & h_{l,v} = ReLU(W_l a_{l,v} + b_l) \label{eq:2}
\end{align}
It has a clear boundary between the aggregation phase (\eq~\ref{eq:1}) and the combination phase (\eq~\ref{eq:2}). As GCN only uses the result after aggregation $a_{l,v}$ in combination phase, \name\ can support GCN with its native events, without a need for user-defined events, as shown in \fig \ref{figure:showcase_configs}(a).
Line 1-4 of the model description file describes the operations in one GCN layer. 
Since \name\ partitions the model configuration by aggregation phase and the events are only created at the end of a partition, the operation in line 1 will be skipped due to an empty event queue. Then a mandatory event creation for changed edges will be conducted after skipping line 1. To begin the next partition, events are grouped and reduced with the aggregation function indicated in line 2. Then each unique target node goes through the process in line 2-5 iteratively. In line 2, the incremental update is applied for the target node with grouped events. In lines 3-5, the updated aggregation result is transformed by the operation indicated in the line. At the end of the partition, each target node propagates its new embedding to neighbors if the embedding changes. Then events for endpoints of all changed edges are created.

\subsection{GraphSAGE}
GraphSAGE\cite{hamilton2017inductive} is proposed to solve the problem of generalization to unseen nodes. It samples the neighborhood and concatenates the sampled neighborhood information with the node embedding.
The mathematical expression of a GraphSAGE layer is:
\begin{align}
    &a_{l,v} = \mathcal{A}(\{h_{l-1,u}: u\in S(v) \}), \label{eq:3} \\
    & h_{l,v} = ReLU(W_1 a_{l,v} + W_2 h_{l-1,v}) \label{eq:4}
\end{align}
$S(v)$ is a sampling of the neighborhood. In GraphSAGE, two different weight matrices are used separately for aggregated neighborhood information and the node itself.  Since the combination phase (\eq~\ref{eq:4}) also uses $h_{l-1,v}$, we pass the changed embedding of a node in one layer to itself in the next layer through user-defined events in {\fontfamily{qcr}\selectfont user\_propagate}. As shown in the model description file in \fig \ref{figure:showcase_configs}(b), after getting 
{\fontfamily{qcr}\selectfont x=}$a_{l,v}$ with the incremental update (line 1), 
we calculate the left side of addition in \eq~\ref{eq:4} $W_1 a_{l,v}$ by executing
{\fontfamily{qcr}\selectfont x=convs[0].lin\_l(x)} (line 2). 
After that, the customized {\fontfamily{qcr}\selectfont user\_apply} function is executed to calculate the right side of addition  (line 3). For a node $v$ in layer $l$, if $h_{l,v}$ doesn't evolve over time, which means it won't receive a user-defined event and $h_{l,v} = h_{l,v}^-$, we can load the value in previous timestamp and do the addition. Otherwise, the updated $h_{l,v}$ can be found in the user-defined event. After applying ReLU (line 4), we can get $h_{l,v} = ReLU(W_1 a_{l,v} + W_2 h_{l-1,v})$. Since \name\ is already memory-friendly and GraphSAGE adopts random sampling with no improvement in model accuracy, we skip the neighborhood sampling process.

\subsection{GIN}
Graph Isomorphism Network (GIN) \cite{gin} is a highly expressive GNN and as powerful as the Weisfeiler-Lehman graph isomorphism test. It contains a weighted aggregation \cite{lightgcn} and uses multilayer perceptrons (MLP) in the combination phase. The mathematical expression of a GIN layer is:
\begin{align}
    & a_{l,v} = \mathcal{A}(\{h_{l-1,u}: u\in N(v) \}), \label{eq:5} \\
    & h_{l,v} = ReLU(MLP((1+\epsilon) h_{l-1,v} + a_{l,v}))\label{eq:6}
\end{align}
Configuring for GIN is similar to but simpler than GraphSAGE, and we will not further explain it here.

\section{Experiment Evaluation}
\begin{table}[t] \footnotesize
\caption{Configuration of GNN models.}
\label{tab:gnn_models}
\centering
\begin{tabular}{||ccccc||} 
 \hline
 Model & \#Layers & Hidden State Len. & Aggr. & Combination \\
 \hline\hline
 GCN\cite{gcn}       & 2 & 256 & min & $W h_v+b$\\ 
 GraphSAGE\cite{hamilton2017inductive}  & 2 & 256 & min & $W_1 a_v + W_2 h_v$\\
 GIN\cite{gin}        & 5 & 64  & max & MLP$((1+\epsilon) h_v + a_v)$\\
 \hline
\end{tabular}
\end{table}
\begin{table}[t] \footnotesize
\caption{Datasets}
\label{tab:datasets}
\centering
\begin{tabular}{||ccccc||} 
 \hline
   Dataset  & \#Vertex & \#Edges & Feature Len. & Max Deg.\\
 \hline\hline
 Cora\cite{citationfull_reddit}   & 19,793 & 126,842 & 8710 & 297 \\ 
 PubMed\cite{planetoid} & 19,717 & 88,648 & 500 & 171\\
 Reddit\cite{citationfull_reddit}  & 232,965 & 13,954,819  & 602 & 21657\\
 Yelp\cite{yelp} & 716,847 & 114,615,892  & 300 & 4886\\
 \hline
\end{tabular}
\end{table}
\subsection{Experiment Setup}
\noindent\textbf{Benchmark GNN models.}
\tab \ref{tab:gnn_models} shows the models for evaluation. We choose the number of layers, hidden state length, and combination process as the examples in PyG library \cite{pyg} and limit the aggregation function to monotonic functions. For all three models, we use ReLU as the activation function. To ensure the effect of changed edges only propagates along the edges or passes to the node itself in the next layer, we disabled normalization, if any.

\noindent\textbf{Benchmark Graph Datasets.}
\tab \ref{tab:datasets} lists benchmark graph datasets commonly used in the GNN domain for social networks\cite{yan2020hygcn}. To simulate streaming graph dynamics, we assign random creation and deletion time to edges as done in T-GCN\cite{huan2022t}. 
Given the significant variation in node degree, we establish a baseline by considering the graph, which has 1,000 updates from the most recent one, as our starting reference. Subsequently, the time span is evenly divided based on the  {\fontfamily{qcr}\selectfont num\_updates}  edges between two timestamps. We use {\fontfamily{qcr}\selectfont num\_updates} values of 1, 10, and 100 to emulate swift real-time updates. Thus, when  {\fontfamily{qcr}\selectfont num\_updates}  is set to 1, we have 1,000 intervals for evaluating execution time. Conversely, with  {\fontfamily{qcr}\selectfont num\_updates}  set to 100, there will be only 10 sample intervals for evaluation.
We use the latest 500k edges from each dataset to capture a graph's snapshot, excluding overly dated interactions that may introduce noise\cite{dygnn}. This approach also avoids memory overloads for accessing the whole $k$-hop neighborhoods of high-degree nodes.
As GIN is initially proposed for graph isomorphism tests and ego networks are used for evaluation\cite{gin}, we follow a classic way \cite{tudataset} to use the one-hot encoding of node degree as the node feature vector.  Due to memory constraints, particularly for Reddit with a large max degree, we cap the one-hot encoding at length 500.

\noindent\textbf{Implementations and Baseline Methods.}
We compare \name\ against two baseline methods: 1) \underline{Full}: inference for the whole graph; 2) \underline{Affected}: inference for the theoretically affected area. Both baseline methods only take the latest snapshot of graph structure as input without knowledge of previous timestamps.
\name\ and baseline methods are all implemented with PyTorch Geometric library\cite{pyg}. As the dataset and consequent intermediate results cannot fit into GPU memory, we use mini-batch inference for the whole graph. We use neighbor loader \cite{hamilton2017inductive} as the mini-batch data loader for both full graph and affected area inference. The neighbor loader returns a re-indexed subgraph of a $k$-hop neighborhood for a set of node indices. For model inference of GIN, we build a customized data loader that generates a directed $k$-hop ego network for each node in a set, with each ego network treated as an independent graph. All data loaders for the baseline experiment use 16 as batch size running on 4 workers. For \name, the intermediate result is directly loaded and accessed on the host, and the model weights are pushed to GPUs at the beginning, if any. During the execution, all unique target nodes in a layer are processed sequentially. The intra-layer incremental update is done on the host as InkStream needs to fetch the intermediate result from host memory, and the dense computation of the combination phase as well as the following activation function is done on the GPU.  Before(after) the incremental update in the aggregation phase, the node embedding will be moved out(to) the GPUs for dense computation of the combination phase.

\begin{table}[t] \scriptsize
\caption{System Configurations.}
\label{tab:sys_configs}
\centering 
\begin{tabular}{||cccc||} 
 \hline
            & CPU &  GPU-N &  GPU-A \\
 \hline\hline
CPU Units   & Intel Xeon 6330 & Intel Xeon 6230  & AMD EPYC 7V13 \\ 
CPU Count & 112 & 104 & 128 \\
Host Memory & 528 GB    & 264 GB    & 528 GB \\ 
GPU Units   & - & NVIDIA A6000 & AMD MI100 \\ 
GPU Count & - & 1 & 1\\
GPU memory  & - & 48.3 GB  & 32 GB \\ 
Off-chip Bandwidth & 5.0 GB/s & 6.0 GB/s  & 9.0 GB/s\\
CPU-GPU Bandwidth & -  & 12.3 GB/s & 26.5 GB/s \\
 \hline
\end{tabular}
\end{table}

\noindent\textbf{Platforms.}
To compare the performance, we evaluate baseline methods and \name\ on three Linux platforms. 1) \underline{CPU}: a CPU cluster with 112 cores; 2) \underline{GPU-N}: a CPU-GPU platform with 104 CPU cores and 1 NVIDIA RTX A6000 GPU;  3) \underline{GPU-A}: a CPU-GPU platform with 128 CPU cores and 1 AMD MI100 GPU.
\tab \ref{tab:sys_configs} lists the system configurations for the above platforms. The off-chip memory bandwidth is evaluated though the MEMCPY mode of \texttt{mbw}\cite{raas2023mbw}. \texttt{mbw} is an in-memory benchmark tool to evaluate the performance of different memory operations, and MEMCPY invokes the memcpy() function to copy a whole data array to another memory area. We use \texttt{nvbandwidth} tool \cite{nvbandwidth} to evaluate the bandwidth from CPU to GPU for GPU-N. On GPU-A, we measure the effective bandwidth between CPU and GPU through the hipMemcpy() function in the HIP library.

\section{Experiment Result}
\subsection{Performance}
\begin{figure}[t] 
\centering
\includegraphics[width=\columnwidth]{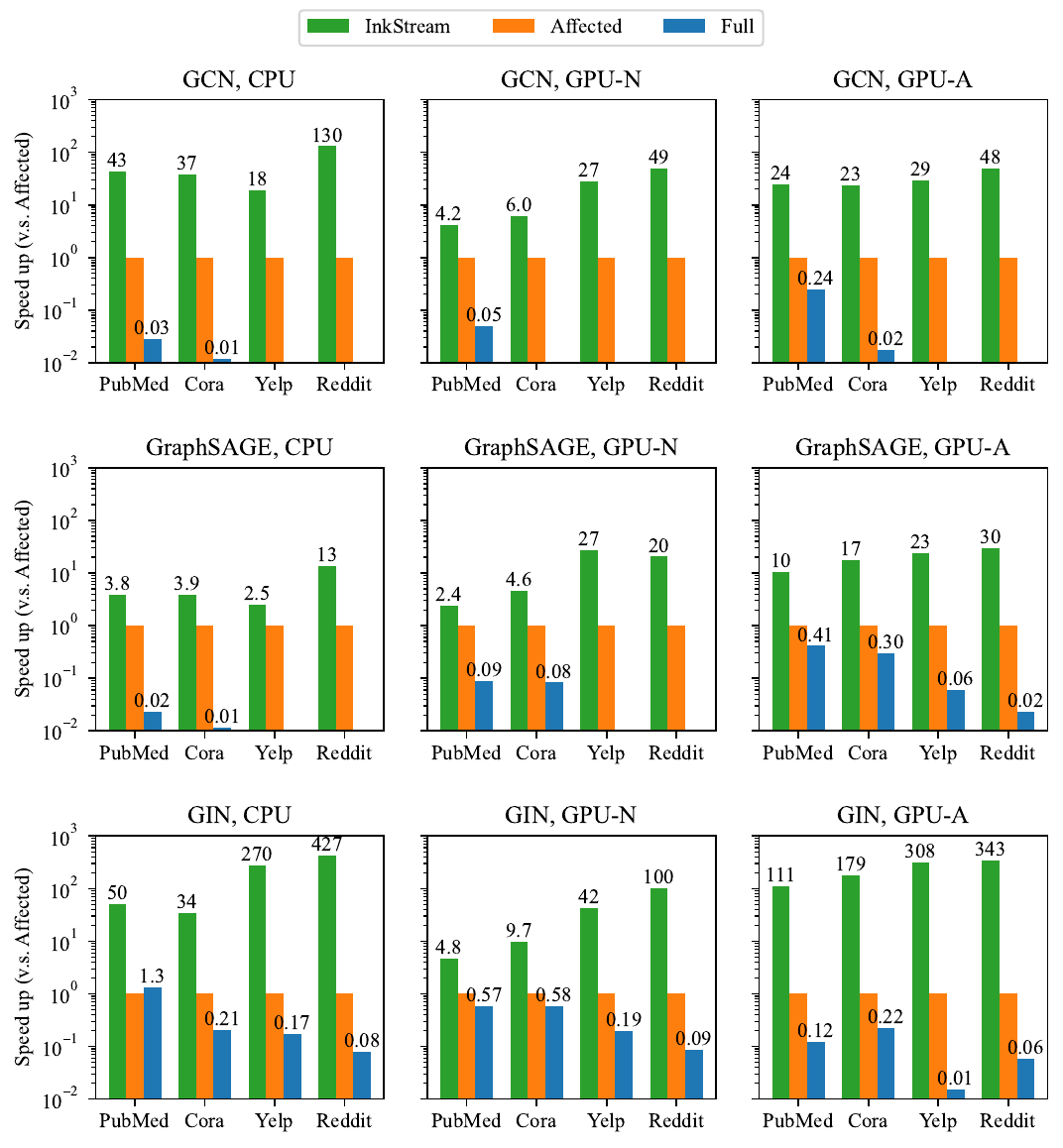}
\caption{Speedup compared with Affected, log scale used for y-axis. Only one edge updates between two timestamps. Missing bars of Full means the speedup is lower than 0.01.}
\label{fig:speedup}
\end{figure}

\fig \ref{fig:speedup} shows the comparison of execution time for three methods, assuming only one edge changed between two timestamps. Especially, we choose the classic method of inference only on the affected area (Affected) as the baseline to compare against. Inkstream shows 2.5-427$\times$ speedup across different platforms. Note that InkStream does not take advantage of parallelization. Meanwhile, classic methods enjoy the multi-processing on 4 workers. InkStream natively supports GCN without additional code. 

Among all three models, GraphSAGE has the lowest speedup. This is because in GraphSAGE, the effect of changed edges propagates both along the graph edges and to the node itself in the next layer, and the change of a node in the previous layer must affect itself in the next layer (refer to \eq~\ref{eq:4}), making the nodes easily impacted. 
Additionally, for platforms with GPU, data used in {\fontfamily{qcr}\selectfont user\_apply} function of GraphSAGE need to be first pushed to the GPU for the computation, causing more CPU-GPU communication.

Across all three platforms, InkStream shows superior speedup on GIN, despite enduring the common difficulty of GraphSAGE. This is because the classic inference methods of the 5-layer GIN take disjoint ego networks as inputs to facilitate the final graph-level pooling layer, and a 5-hops ego network could reach 80\% of the whole graph, as previously shown in \fig\ref{figure:motivating_data}(a). This means that most nodes have fetched the same area of the graph and done repeated computations. More importantly, the input feature vector (limited to the length of 500) has a nearly 10$\times$ size compared with the final node embedding (length of 64). By managing the large ego networks with heavy feature vectors, classic methods suffer from tremendous memory access and repeated computation. However, in our method, we update the node embeddings first without isolating each ego network, so there is no such repeated computation. Then the pooling operation is only done after the update, taking the short node embeddings as input, thus largely reducing the memory access and computation.

\begin{figure}[t] 
\centering
\includegraphics[width=\columnwidth]{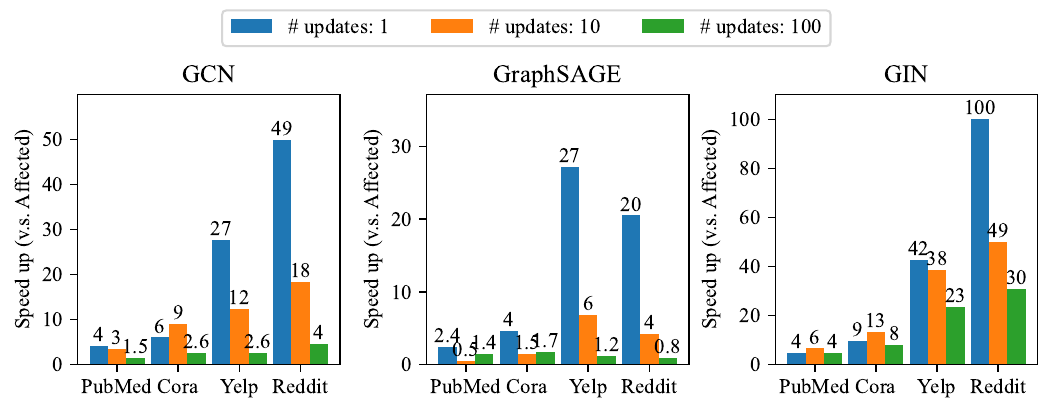}
\caption{The effect of the number of updates on the speedup, evaluated on GPU-N.}
\label{fig:speedup_batch}
\end{figure}


\noindent \textbf{Impact of Number of Updates.}
\fig \ref{fig:speedup_batch} shows the speedup against the classic method of affected area inference with the different number of updates between two timestamps. Regardless of the model, the speedup shows a trend of decreasing when the number of updates increases. This is because when the number of updates increases, one node could have more nodes in its neighborhood to be changed, thus bringing more chance to be affected. Meanwhile, InkStream monitors the changes in nodes to avoid unnecessary computation and memory access. However, when the affected area is too large and a non-trivial portion of the nodes change, the saving brought by computation skipping will not be enough to offset the overhead of monitoring the changes in nodes. This is similar to the situation that sparse methods may perform worse than a simple dense method when the data is not sparse enough. 

\subsection{Applicable Condition Distribution}
\begin{figure}[t] 
\centering
\includegraphics[width=\columnwidth]{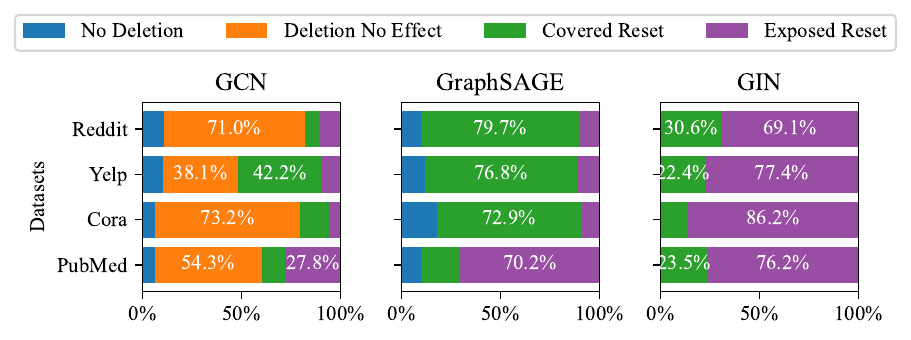}
\caption{Distribution of conditions of events' effect. A sum of the conditions occurrences in different layers across multiple runs is used to calculate the distribution. For conditions of "no deletion", "deletion (has) no effect", and "covered reset", the node receiving events can be incrementally updated. Otherwise, recomputation is required to ensure a correct result.}
\label{fig:cond_dist}
\end{figure}

If a node is applicable with the incremental update, we can bypass accessing the whole neighborhood from memory and use the newly generated events in the previous layer to compute. \fig \ref{fig:cond_dist} shows the distribution of potential effects of events on a node. In GCN and GraphSAGE, over 70\% of the visited nodes can be incrementally updated. In GIN, the exposed reset takes a non-negligible portion. This is because GIN has a deeper network meanwhile suffers from sensitive node embedding due to the self-impact. In the last layer, one node could have most of its neighborhood affected, causing unrecoverable data loss.

\subsection{Memory Access}
\begin{figure}[t] 
\centering
\includegraphics[width=\columnwidth, trim={0 0.1cm 0 0.1cm},clip]{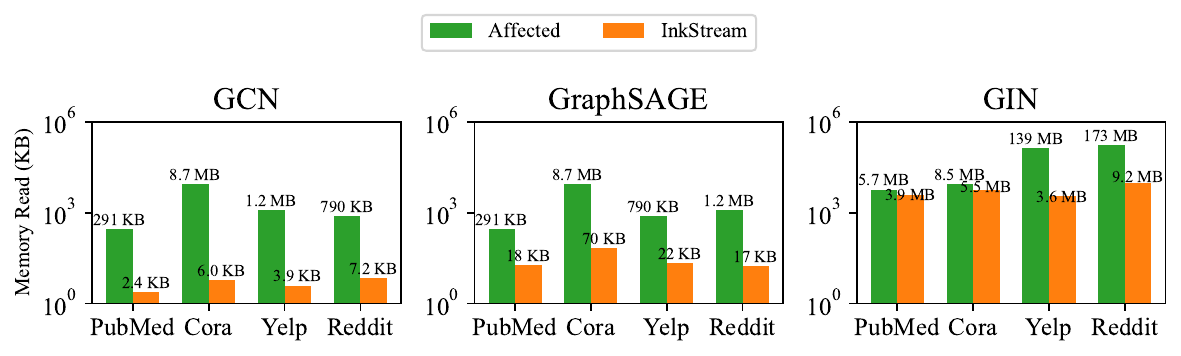}
\caption{Memory access assuming only one edge update. Log scale is used for y axis.}
\label{fig:memory_access}
\end{figure}
We evaluate the amount of memory access by counting the number of nodes whose input feature is read in the Affected method. For InkStream, we count the number of times accessing the saved intermediate result of node embeddings. Besides the reduced memory access brought by incremental updates, InkStream also naturally benefits from the short length of intermediate node embeddings. Especially in GCN, the memory access is reduced by 110-1485$\times$. This is because the GCN layer has the combination phase executed before the aggregation phase. So in the first layer, InkStream doesn't need to access the long input feature vector. Note that in the first layer, the direct input of the aggregation phase is the input of the model, the feature vector, if the aggregation phase is first executed.
For GraphSAGE and GIN, the node embedding is more sensitive, leading to more influenced nodes, and the aggregation phase is executed after the combination phase. So in the first layer, InkStream needs to fetch the heavy node feature. Despite these demerits, with a shallow network of GraphSAGE, InkStream still achieves a 16-1273$\times$ reduction in memory access. However, when the model is deeper and a non-trivial portion of nodes are affected, InkStream suffers from the high recompute rate ("Exposed Reset" in \fig \ref{fig:cond_dist} ), thus cannot bring significant improvement. The reduction in memory access is only 1.5-18.8$\times$.


\subsection{Additional Memory Cost}
\label{sec:memory_cost}
In this section, we analyze the additional memory cost for saving intermediate results of a previous timestamp. As mentioned previously, to avoid the over-smoothing problem, GNNs are usually designed shallow ($k$=2 for most cases), with a carefully tuned bandwidth \cite{yoo2023sgcn, gcn, gin, hamilton2017inductive, gat, chiang2019cluster, dgcnn, zhang2021graph, yelp, schlichtkrull2018modeling}. Compared with the input feature vector (8710 in Cora), the intermediate result of node embedding is light (usually set to 16/32/64/128/256). Moreover, in one layer, InkStream only sets two checkpoints for saving the intermediate result: immediately before and after the aggregation phase. Therefore, the saved intermediate result will not bring exploding memory costs. In our experiments, the additional memory overhead is 0.12-3.4$\times$ the size of the dataset for GIN and SAGE. For GIN, the overhead is 1.3$\times$ the dataset, supposing one-hot encoding of degree for feature vector, with an upper limit of 500.

\section{Related Works}

\subsection{Streaming Graph Neural Networks}
Most works \cite{pareja2020evolvegcn, he2023dynamically} focus on the rich time information in streaming graphs and provide an evolving model with consistently high accuracy across varying timestamps. However, these methods have non-trivial retraining overhead on the whole graph and are unsuitable for real-time updating.
DyGNN \cite{dygnn} is one of the works that requires little overhead by only updating the edge endpoints and their immediate neighbors with customized GNN units separately. Despite the little overhead, DyGNN only provides a solution for edge insertion and is incompatible with GNNs with more than 2 layers.  
 T-GCN \cite{huan2022t} boosts the retraining process for streaming graphs by accelerating the time-consuming sampling process through efficient sampling methods and locality-aware data partitioning for multi-GPUs. Although the acceleration brings 10-100x speedup, considering the substantial overhead of retraining, it still takes minutes to hours, which is not applicable in real-time scenarios. AEGNN \cite{schaefer2022aegnn} targets object classification tasks by transforming images into point clouds and processing a video as an evolving spatio-temporal graph. Although AEGNN is also memory efficient by only processing the affected area of "difference" between two successive images, the graph structure remains static over time.

\subsection{Streaming Graph Analytics}
\label{sec:streaming_graph_analytics}
Streaming graph analytics refers to executing traditional graph processing algorithms on streaming graphs, e.g. ShortestPath, PageRank. It has been extensively studied and explored \cite{commongraph, kickstarter, graphbolt, tdgraph, dzig, naiad, kineograph, jetstream}. Each time, a batch of edge changes is applied to the graph, and the vertex state needs to be reevaluated, like distance in ShortestPath and importance in PageRank. Given that edge modifications are typically tiny compared to the graph's size, an incremental update approach is favored over recalculating from scratch. The operations performed for incremental updates vary depending on the algorithm type (\textit{accumulative} and \textit{monotonic}). 
\\
\textbf{Accumulative Algorithms:}
 These, like PageRank, update the vertex state through accumulative operations such as summation. The incremental update process first negates the previous state's influence and then propagates the new one. Beginning with the changed edges' endpoints, affected nodes in each iteration counteract the influence of the prior state and then disseminate the new state.
\\
\textbf{Monotonic Algorithms:}
Monotonic algorithms (e.g., ShortestPath and ConnectedComponents) update the vertex attributes with a selective operation (e.g., min, max). For an inserted edge $(u,v)$, the destination vertex $v$'s new state is calculated, followed by the propagation of the new value by affected vertices. Edge deletion is more complex, involving state recovery and incremental convergence \cite{kickstarter, jetstream}. In the recovery phase, affected vertices are identified and reset. During incremental convergence, these vertices assimilate their neighbors' states and then propagate their new values until all vertices stabilize.

However, the streaming graph analytics methods cannot be directly applied to GNNs due to 3 major differences.
\begin{enumerate}
\item  \textit{Affected Area for Changed Edges}. In traditional algorithms, a modified edge's influence can span distant nodes. In contrast, in the $k$-layer GNN model, the changed edge can influence up to $k$ hops from its endpoints in a $k$-layer GNN.
\item  \textit{Computation Data Size}.
Traditional algorithms often employ a scalar or a short vector for the vertex state. Conversely, GNNs' hidden states can range from tens to thousands in length, posing storage and computation challenges.
\item \textit{Spatial and Temporal Transitivity}.
Traditional algorithms frequently exhibit both spatial and temporal transitivity in node states. Spatially, a node can only have a shorter distance than its descendants (for instance, in SSSP). Temporally, with edge insertions, node distances can only be non-increasing. Contrarily, GNN layers do not exhibit this kind of transitivity due to their non-monotonic combination functions. In traditional algorithms where a node's state might consistently improve, GNNs require canceling previous impact first.
\end{enumerate}

\section{Limitation and Future Work}
Currently, our method can only support networks with min/max as the aggregation function. However, \name\ can be easily extended to accumulative aggregation functions, e.g., incrementally updating by adding up the delta value, similar to the delta-accumulative incremental computation model for graph processing \cite{zhang2013maiter}. It is worth noting that, affected by the expressiveness of floating point numbers, there will be a loss in the delta-accumulative method, gradually increasing with the number of layers. Moreover, we haven't optimized the communication between the host and GPUs nor increased the parallelism (single thread, single worker), making \name\ reach a limited speedup on multi-GPU clusters. However, it is possible to parallelize the processing of nodes in one layer. Finally, \name\ is now only designed for node embedding calculation and does not support global pooling layers. 

\section{Conclusion}
This paper studies GNNs on streaming graphs, uncovering a notable difference between theoretical and actual affected nodes for models using monotonic aggregators. Building on this insight, this paper presents \name, an innovative approach that trims redundant computations and reuses the calculated embeddings in the previous timestamp for incremental updates, achieving remarkable performance boosts without compromising accuracy. 
The experiment results show that InkStream accelerates GNN inference by 2.5-427$\times$ on a CPU cluster and 2.4-343$\times$ on two different GPU clusters.



\bibliographystyle{plain}
\bibliography{references}

\end{document}